\documentclass{article}

\usepackage{xurl}
\usepackage{microtype}
\usepackage{xcolor}
\usepackage{graphicx}
\usepackage{subfigure}
\usepackage{booktabs} 
\usepackage{subcaption} 
\usepackage{hyperref}

\usepackage[accepted]{mlsys2025}

\mlsystitlerunning{OEA: Batch-Aware Expert Routing for Faster Decode Without Retraining}

\newcommand{\methodname}{OEA}

\usepackage{multirow}

\usepackage{amsmath,amsfonts,amsthm,bm,bbm,amssymb}
\newcommand{\R}{\mathbb{R}}
\def\vx{{\bm{x}}}
\newcommand{\indicator}{\mathbbm{1}}
\newcommand{\E}{\mathop{{}\mathbb{E}}}
\newcommand{\se}{\text{se}}
\begin{document}

\twocolumn[
\mlsystitle{Opportunistic Expert Activation: Batch-Aware Expert Routing for Faster Decode Without Retraining}



\mlsyssetsymbol{equal}{*}

\begin{mlsysauthorlist}
\mlsysauthor{Costin-Andrei Oncescu}{harvard,together}
\mlsysauthor{Qingyang Wu}{together}
\mlsysauthor{Wai Tong Chung}{together}
\mlsysauthor{Robert Wu}{together}
\mlsysauthor{Bryan Gopal}{together}
\mlsysauthor{Junxiong Wang}{together}
\mlsysauthor{Tri Dao}{princeton,together}
\mlsysauthor{Ben Athiwaratkun}{together}
\end{mlsysauthorlist}

\mlsysaffiliation{harvard}{Harvard University. Part of the work was done when Costin was interning at Together AI.}
\mlsysaffiliation{together}{Together AI}
\mlsysaffiliation{princeton}{Princeton University}

\mlsyscorrespondingauthor{Costin-Andrei Oncescu}{concescu@g.harvard.edu}

\mlsyskeywords{Machine Learning, MLSys, LLMs, Efficient Decoding, Mixture Of Experts}

\vskip 0.3in

\begin{abstract}
An increasing number of LLMs employ Mixture-of-Experts (MoE) architectures where the feed-forward layer is replaced by a pool of experts and each token only activates a small subset of them. During autoregressive generation, these models often enter a memory-bound regime even for moderate batch sizes because the average expert load grows more slowly than in an equivalent dense feedforward layer. Consequently, MoE latency is governed by the number of activated experts. We introduce a framework for \textbf{dynamically} re-routing token-to-expert mapping to lower this number (and thus, the decode latency) while preserving a comparable quality. Our best results use a \textbf{batch-aware routing} that works by having tokens \textbf{piggyback} experts that have already been loaded into memory due to being crucial to other tokens within the same batch. Empirically, we evaluate our method on the Qwen3-30B and Qwen3-235B models with a batch size of $16$. Without any statistically significant loss in accuracy, our approach achieves latency reductions of $39\%$ and $15\%$ in the MoE layer decode latency, respectively.
\end{abstract}
]



\printAffiliationsAndNotice{}  

\section{Introduction}
\label{sec:intro}
Mixture-of-Experts (MoE) architectures have contributed significantly to the state-of-the-art in language modeling \cite{liu2024deepseek, team2025kimi, yang2025qwen3}. They replace the feedforward layer with a pool of experts --- smaller feedforward layers --- and route each input to only a small subset of the pool. By employing this sparse and conditional computation, MoEs decouple model size from the computation cost, allowing for more amenable model scaling.

When deploying these models, serving frameworks~\cite{kwon2023efficientVLLM,zheng2024sglang} usually batch several requests and proceed in two steps: prefill and decode. During the prefill stage, prompts are processed together in parallel across sequence length, much as a normal forward pass would. Then, decoding is the process of sequentially (autoregressively) generating one new token at a time, in parallel across a batch. This stage has a lower arithmetic intensity than prefill and is often memory-bound~\cite{rajbhandari2022deepspeedMoE}, where runtime is limited by data movement bandwidth rather than arithmetic throughput.

Because decoding dominates serving time for long sequences and interactive workloads, reducing its latency directly improves user experience and cost efficiency.

\textbf{The problem.} During decode, it takes a larger batch size to get into a regime where experts are not memory-bound. This is because when each token activates $k$ experts out of $N$, the average per-expert load increases only at a rate of $k/N$ which is low by design in MoEs (e.g. $1/16$ in Qwen3). Coupled with the arithmetic intensity being roughly 100-200~\cite{nvidia2022h100WhitePaper}, the sparsity factor $N/k$ results in required batch sizes of order of thousands for MoEs to be in compute bound regime (e.g. $\approx 1.6k$ for Qwen3).
Hence, for moderate batch sizes, the latency of an MoE layer is not dominated by the computational load of individual experts, but rather by the overhead of fetching the weights of all activated experts from the high-bandwidth memory (HBM) to the on-chip one (SRAM)~\cite{rajbhandari2022deepspeedMoE}. Consequently, latency becomes effectively linear in the number of unique activated experts, a number that can grow quickly with batch size in spite of each token activating only a few experts; this is because we need to activate the \emph{union} of all these small sets of experts.

This paper introduces Opportunistic Expert Activation (\methodname), a batch-aware routing framework designed to lower decode latency by explicitly minimizing the number of unique active experts per batch during inference. \methodname\ operates without any model retraining and comprises two stages:

\begin{algorithm}[tb]
   \caption{Simplified \methodname\ Routing Algorithm}
   \label{alg:simplified-router}
\begin{algorithmic}[1]
   \STATE {\bfseries Input:} Token embeddings $\vx_{1..B}$, Initial number of experts per token $k$, Sorted expert indices $e_{i,j}$ for each token $i$ and rank $j$. Hyperparameter: $k_0$.
   \STATE \COMMENT{Phase 1: Determine Baseline Experts}
   \FOR{$i=1$ {\bfseries to} $B$}
      \STATE $S^{\text{base}}_i \leftarrow \{e_{i,1}, \dots, e_{i,k_0}\}$
   \ENDFOR
   \STATE
   \STATE \COMMENT{Phase 2: Opportunistic Piggybacking}
   \STATE $S^\text{base} \leftarrow \bigcup_{i=1}^B S^{\text{base}}_i$ \COMMENT{Union of all required experts}
   \FOR{$i=1$ {\bfseries to} $B$}
      \STATE $S_i \leftarrow S^{\text{base}}_i$ \COMMENT{Initialize final set with baseline}
      \FOR{$j=k_0+1$ {\bfseries to} $N$}
         \IF{$|S_i| > k$}
            \STATE {\bfseries break}
         \ENDIF
         \IF{$e_{i,j} \in S^\text{base}$}
            \STATE $S_i \leftarrow S_i \cup \{e_{i,j}\}$
         \ENDIF
      \ENDFOR
   \ENDFOR
   \STATE {\bfseries Output:} Final expert sets $S_1, \dots, S_B$
\end{algorithmic}
\end{algorithm}
\begin{enumerate}
\item Firstly, \methodname~sets a minimum quality baseline for each token by keeping the first few of its expert choices to guarantee crucial computations take place.
\item Then, \methodname~opportunistically augments this baseline by routing tokens to additional, lower-priority experts only if those experts already need to be loaded due to another token’s baseline requirement within the same batch.
\end{enumerate}
This ``piggybacking" mechanism allows the model to recover some of the performance that is potentially lost due to activating fewer experts, practically for free since it preserves the number of activated experts (and thus latency).

\textbf{Relation to Prior Work.} \methodname\ is complementary to approaches that reduce the number of experts activated \emph{per token}~\cite{lu2024notAllExpertsEq}. In contrast, our piggybacking phase can be applied on top of such methods at no added cost. Moreover, unlike prior dynamic batch-aware routing strategies~\cite{gupta2024lynx}, \methodname\ guarantees a batch-independent quality baseline for every token, ensuring consistent per-token computation regardless of batch composition.

\textbf{Contributions} Our contributions are as follows:
\begin{itemize}
    \item We formalize the MoE decode latency problem under a memory-bound roofline model, showing that reducing the number of unique active experts is the primary optimization target.
    \item We propose \methodname, a dynamic routing algorithm that provides a tunable trade-off between model quality and system performance.
    \item We evaluate \methodname\ on the Qwen3-30B and Qwen3-235B models, demonstrating its ability to substantially reduce the number of active experts and, consequently, MoE latency, while maintaining performance on both downstream tasks and language modeling perplexity. At a batch size of $16$, \methodname\ achieves latency reductions of $39$\% on the 30B model and $15$\% on the 235B model.
\end{itemize}

\section{Background and Motivation}
\label{sec:backgroundMotivation}
\begin{figure}[t]
  \centering
  \includegraphics[width=.9\linewidth]{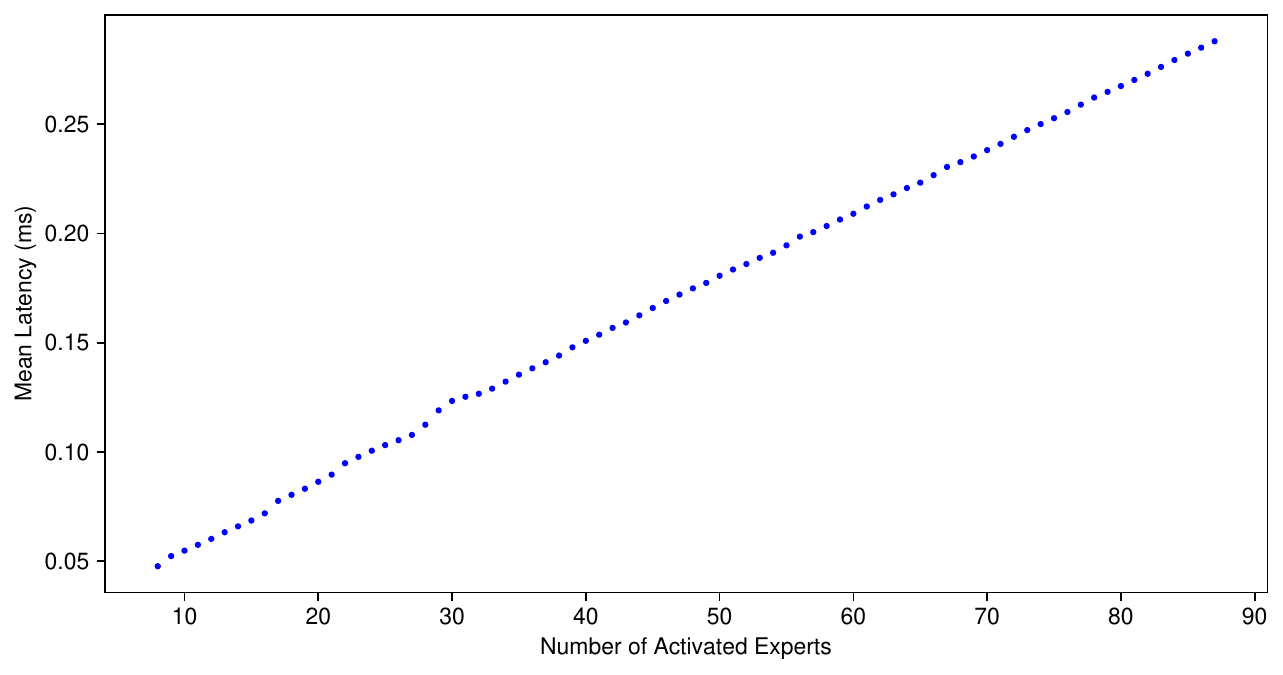}
  \caption{Mean MoE latency as a function of the number of activated experts within a decode batch. The average is computed over all layers and decode steps across a GPQA evaluation of the vanilla Qwen3-30B-A3B model.}
  \label{fig:expert_latency_gpqa}
\end{figure}
Modern state-of-the art MoE models such as Kimi K2~\cite{team2025kimi}, Deepseek-V3~\cite{liu2024deepseek} and Qwen3~\cite{yang2025qwen3} fundamentally incorporate the same setup (excluding potentially shared experts) popularized by Shazeer et al.~\yrcite{shazeer2017outrageouslyFirstMoE} --- namely, they replace the feedforward layer of a transformer with sets of $N$ experts $E_1,\ldots E_N : \R^D \to \R^D$ (where $D$ is the embedding dimension) and a router scoring function $R : \R^D \to \Delta^N$ that assigns a normalized score $R(\vx)_i$ to each expert $i$. The output of the MoE module is then computed via:
\begin{align}
\label{eq:moeDefinition}
    \text{moe}(\vx)=\sum_{i \in S} \frac{R(\vx)_i}{\sum_{j\in S}R(\vx)_j}E_i(\vx)
\end{align}
where $S=\text{Top}_k(R(\vx))$ is the set of indices of top-$k$ values of the router's scores. The extra normalization factor is optional, but enabled in Qwen3, the model we evaluate.
Henceforth, we use $B$ for batch size.

Typically, serving these models is done by batching requests and proceeding in the following two stages:
\begin{enumerate}
    \item The \textbf{prefill stage}, where activations and KV caches are computed for the prompts. This passes over entire prompts' at once, thus increasing the effective (token) batch size (\textit{i.e.}, sequence length $\times$ batch size) of the MoE layers, resulting in heavier loads for each expert.
    \item The (iterative) \textbf{decode stage} where, at a given decode step, exactly one token of each sequence in the batch is processed for next-token prediction. Crucially, the effective batch size passed to the MoE layers is now only equal to the batch size.
\end{enumerate}

Note that the low effective (token) batch size seen during decoding is further exacerbated by the fact that the average expert load only increases at a rate of $k/N$ per token. This raises the threshold batch size for reaching the compute-bound regime, implying that even moderately-sized batches can still result in memory-bounded experts. In this regime, the time to fetch expert's weights from HBM into on-chip SRAM dominates the time needed to compute $E_i$'s outputs. 
Consequently, for each expert, latency depends primarily on whether it is activated at all: if no token is routed to it, its weights need not be fetched and thus incur no latency, whereas once it is activated and fetched, the marginal cost to serving additional tokens is negligible. Therefore, when experts are not executed in parallel, overall MoE latency scales with the number of activated experts.

To illustrate how quickly this quantity grows, consider, for example, a setting where each token activates $k=8$ experts out of $N=128$ total ones (as is the case for Qwen3). For a batch size of $16$ tokens, any number of experts between $8$ and $128$ could be employed. Assuming uniform routing (which the models are trained to balance), the expected number of activated experts is $82$\footnote{The exact formula is $N(1-(1-\frac{k}{N})^B)$ where $B$ is the batch size.}. Note that this represents an increase of up to $10\times$ over a batch size of $1$ (where the token only triggers $k=8$ experts). This is not the case in non-MoE architectures where both a batch size of $1$ and one of $16$ would deem a memory-bound regime and thus incur a fixed one-time fetching cost.

In summary, reducing the number of activated experts directly targets the dominant term of MoE decode latency in the memory-bound regime.
We cover the rest of the related work in Section~\ref{sec:related-work} where we also put our method in perspective.
\section{Our framework}
\begin{table*}[t]
\caption{Ablation across $k_0$: Benchmark accuracies for Phase 1 (pruned, top-$k_0$) vs simplified OEA routing (top-$k_0$+piggybacking) on Qwen3-30B-A3B. \textbf{Pruned} refers to using top $k_0$ experts per token, \textbf{OEA} does additional piggybacking and vanilla represents the default model. Results averaged over $4$ runs. Setups that are no worse than vanilla (standard-error adjusted) are in bold.}
\label{tab:30b-accuracies-merged-bold}
\vskip 0.15in
\begin{center}
\begin{small}
\begin{sc}
\begin{tabular}{lccccccccccc}
\toprule
 & \multicolumn{2}{c}{$k_0=3$} & \multicolumn{2}{c}{$k_0=4$} & \multicolumn{2}{c}{$k_0=5$} & \multicolumn{2}{c}{$k_0=6$} & \multicolumn{2}{c}{$k_0=7$} & \multirow{2}{*}{vanilla} \\
\cmidrule(lr){2-3}
\cmidrule(lr){4-5}
\cmidrule(lr){6-7}
\cmidrule(lr){8-9}
\cmidrule(lr){10-11}
Benchmark & pruned & OEA & pruned & OEA & pruned & OEA & pruned & OEA & pruned & OEA \\
\midrule
aime24 & 51.2 & \textbf{80.0} & 75.8 & \textbf{81.9} & \textbf{80.6} & \textbf{81.5} & \textbf{80.2} & \textbf{80.8} & \textbf{82.5} & \textbf{78.5} & \textbf{80.4} \\
gpqa & 45.7 & \textbf{58.6} & 54.3 & \textbf{59.3} & 56.2 & \textbf{61.1} & 58.3 & \textbf{62.2} & \textbf{59.7} & \textbf{60.6} & \textbf{60.2} \\
livecodebench & 37.4 & \textbf{61.2} & 58.2 & \textbf{62.7} & \textbf{63.2} & \textbf{62.0} & \textbf{63.1} & \textbf{63.1} & \textbf{63.0} & \textbf{62.5} & \textbf{62.1} \\
math\_500 & 91.1 & \textbf{93.5} & \textbf{92.7} & \textbf{93.1} & \textbf{92.6} & \textbf{93.3} & \textbf{93.1} & \textbf{93.1} & \textbf{93.3} & \textbf{93.2} & \textbf{92.8} \\
\bottomrule
\end{tabular}
\end{sc}
\end{small}
\end{center}
\vskip -0.1in
\end{table*}
\begin{table*}[t]
\caption{Ablation across $k_0$: Benchmark accuracies for Phase 1 (pruned, top-$k_0$) and simplified OEA routing (top-$k_0$+piggybacking) on Qwen3-235B-A22B. \textbf{Pruned} refers to using top $k_0$ experts per token, \textbf{OEA} does additional piggybacking and vanilla represents the default model. Results averaged over $3$ runs. Setups that are no worse than vanilla (standard-error adjusted) are in bold.}
\label{tab:235b-accuracies-merged-bold}
\vskip 0.15in
\begin{center}
\begin{small}
\begin{sc}
\begin{tabular}{lccccccccc}
\toprule
 & \multicolumn{2}{c}{$k_0=3$} & \multicolumn{2}{c}{$k_0=4$} & \multicolumn{2}{c}{$k_0=5$} & \multicolumn{2}{c}{$k_0=6$} & \multirow{2}{*}{vanilla} \\
\cmidrule(lr){2-3}
\cmidrule(lr){4-5}
\cmidrule(lr){6-7}
\cmidrule(lr){8-9}
Benchmark & pruned & OEA & pruned & OEA & pruned & OEA & pruned & OEA \\
\midrule
aime24 & 17.5 & 81.4 & 69.4 & 82.5 & 81.9 & \textbf{83.6} & 82.8 & \textbf{83.6} & \textbf{85.0} \\
gpqa & 43.8 & 66.3 & 56.4 & \textbf{67.7} & 60.6 & \textbf{67.5} & 64.1 & 67.5 & \textbf{68.4} \\
livecodebench & 5.7 & 63.4 & 27.4 & 67.1 & 53.5 & 66.1 & 60.8 & 66.1 & \textbf{68.5} \\
math\_500 & 80.9 & \textbf{94.4} & 93.3 & \textbf{94.8} & \textbf{94.5} & \textbf{94.7} & \textbf{94.5} & 94.3 & \textbf{94.7} \\
\bottomrule
\end{tabular}
\end{sc}
\end{small}
\end{center}
\vskip -0.1in
\end{table*}

\subsection{Latency and Number of Activated Experts}
\label{sec:goal-rephrasing}
To formalize the argument introduced in Section~\ref{sec:backgroundMotivation}, we adopt a simplified latency model for the computation performed by one expert. Let $f(n)$ represent the time it takes an expert to process $n$ tokens and let it be given by $f(0)=0$ and $f(n)=an+b$ for $n>0$. Here, $b$ is the cost of fetching the expert's weights from the high-bandwidth memory (HBM) into on-chip SRAM, while $a$ is the computation time it takes to process one token. It follows that the total latency of a whole MoE block is given by:
\begin{align}
\label{eq:moeLatency}
\sum_{i=1}^N f(\text{cnt}_i)
    &=  \sum_{i=1}^N {b\cdot\indicator_{\text{cnt}_i > 0}+a\cdot  \text{cnt}_i} \nonumber\\
    &= b\cdot T + a\cdot Bk
\end{align}
where $\text{cnt}_i$ is the number of tokens routed to expert $E_i$; $T$ is the number of experts that have at least one token routed to them; $B$ is the batch size; $N$ is the total number of experts; and $k$ is the number of experts activated per token.

Equation~\ref{eq:moeLatency} shows that the overall latency is given by a memory-bound term linear in the number of active experts $T$ and a compute-bound term linear in the total computation load $Bk$. Whether we are in a compute- or memory-bound regime only indicates which of these terms dominates, but as a general statement, it directly follows that reducing $T$ lowers the latency. If the loads $\text{cnt}_i$ are small enough to be in a memory-bound regime~\cite{rajbhandari2022deepspeedMoE}, the total latency is dominated by $b \cdot T$ and thus we can expect almost proportional gains to the drop in $T$.

While this description is a simplification --- it does not account for system-level effects such as kernel launch overhead, padding to equalize expert loads, or the use of optimized kernels like Grouped GEMM~\cite{Hejazi2024_GroupedGEMM} --- these factors do not alter the constraint. Grouped GEMM can improve efficiency by batching computations for different experts, but it still requires all activated expert weights to be loaded into on-chip memory, meaning the latency remains fundamentally tied to $T$ in the memory-bound regime. We confirm this empirically (Figure~\ref{fig:expert_latency_gpqa}).

Finally, note that $\text{cnt}_i \leq B$, since each token can route to an expert at most once; this holds for any potential re-routing as well. Furthermore, for the original top-$k$ routing, if we are to further assume it to be uniform, it follows that $\E[\text{cnt}_i]=Bk/N$ which is much lower and thus increases the threshold for $B$ to be in a compute-bound regime. We henceforth turn our focus on optimizing $T$ and assume we are in a regime where this translates (as shown empirically) to lower overall latency.

To achieve this, we modify token routing during inference while preserving empirical performance. This approach is motivated by recent studies demonstrating the robustness of MoE models to re-routing \cite{li2025r2multimodalReroute,gupta2024lynx}. While several approaches have explored \emph{static expert pruning}~\cite{lu2024notAllExpertsEq, liu2024evolutionaryStaticPruning} --- permanently removing experts to save memory --- this inevitably constrains the model's capacity. In contrast, our goal is to develop methods that maintain a minimum level of performance in the worst case while enabling full recovery of the model's original performance in the best case.
\subsection{The Proposed Routing Algorithm}
\label{sec:algorithm}
\textbf{Why two algorithms?} While Algorithm~\ref{alg:router} describes our method --- \methodname\ --- in its \emph{full generality}, following exhaustive experiments, we conclude that a \emph{simplified} version of it, described in Algorithm~\ref{alg:simplified-router}, recovers its performance while requiring fewer hyperparameters. We hereby describe its most general form and then touch on how to simplify at the end of Section~\ref{sec:experiments-ppl}.

Following Section~\ref{sec:goal-rephrasing}'s argument, \methodname\ aims to minimize the number of activated experts $T$ within a decode batch. Its core constraint is to ensure that the overall response quality, for any given sequence in the batch, does not significantly degrade. This motivates a two-stage approach that works by first establishing a quality baseline for each token \emph{independently}, and then opportunistically recovering lost performance by exploiting the shared computation within the batch. 

\begin{algorithm}[tb]
   \caption{\methodname\ Routing Algorithm}
   \label{alg:router}
\begin{algorithmic}[1]
   \STATE {\bfseries Input:} Token embeddings $\vx_{1..B}$, Router scores $R(\vx_i)$, Sorted expert indices $e_{i,j}$ for each token $i$ and rank $j$. Hyperparameters: $k_0, p, k^\text{max}, \text{maxP}$.
   \STATE \COMMENT{Phase 1: Determine Baseline Experts}
   \FOR{$i=1$ {\bfseries to} $B$}
      \STATE Find $t_i = \min \{t' \mid \sum_{j=1}^{t'} R(\vx_i)_{e_{i,j}} \geq p\}$
      \STATE $n_i \leftarrow \min(k_0, t_i)$ \COMMENT{Number of baseline experts}
      \STATE $S^{\text{base}}_i \leftarrow \{e_{i,1}, \dots, e_{i,n_i}\}$
   \ENDFOR
   \STATE
   \STATE \COMMENT{Phase 2: Opportunistic Piggybacking}
   \STATE $S^\text{base} \leftarrow \bigcup_{i=1}^B S^{\text{base}}_i$ \COMMENT{Union of all required experts}
   \FOR{$i=1$ {\bfseries to} $B$}
      \STATE $S_i \leftarrow S^{\text{base}}_i$ \COMMENT{Initialize final set with baseline}
      \FOR{$j=n_i+1$ {\bfseries to} $\text{maxP}$}
         \IF{$|S_i| > k^\text{max}$}
            \STATE {\bfseries break}
         \ENDIF
         \IF{$e_{i,j} \in S^\text{base}$}
            \STATE $S_i \leftarrow S_i \cup \{e_{i,j}\}$
         \ENDIF
      \ENDFOR
   \ENDFOR
   \STATE {\bfseries Output:} Final expert sets $S_1, \dots, S_B$
\end{algorithmic}
\end{algorithm}

\textbf{Notation.} Suppose the $B$ tokens in the batch are $\vx_1\ldots\vx_B$, and that their sorted expert index scores are $e_{i, j}$ where $e_i$ is a permutation such that,
\begin{align*}
    R(\vx_i)_{e_{i, 1}} \geq R(\vx_i)_{e_{i, 2}} \geq \cdots \geq R(\vx_i)_{e_{i, N}},
\end{align*}
where $e_{i, j}$ represents the $j$\textsuperscript{th} expert choice of $i$\textsuperscript{th} token. In particular, the default router (as described in Section~\ref{sec:backgroundMotivation}) routes token $\vx_i$ to experts in the set $\text{Top}_k(R(\vx_i))=\{e_{i, 1}, \ldots, e_{i, k}\}$.
Our target is to decide sets $S_1,\ldots, S_B \subseteq \{1,\ldots N\}$ where $S_i$ represents the set of experts that the $i$\textsuperscript{th} token routes to.

\paragraph{Phase 1: Baseline expert selection.}
The first phase guarantees a minimum foundation for each token, irrespective of how it is batched. For each token $\vx_i$, we create this baseline by activating the first $n_i$ experts, thus creating a base set of experts $S^{\text{base}}_i=\{e_{i, 1}, \ldots, e_{i, n_i}\}$.  This is motivated by empirical findings that the top-ranked experts are disproportionately critical to output quality \cite{gupta2024lynx}. The number of base experts is determined by two hyperparameters: (1) a fixed upper bound $k_0\in\{1\ldots N\}$; and (2) a cumulative score $p\in (0,1]$, following $n_i=\min(k_0, t_i)$ where $t_i$ is the minimum number of experts it takes to reach a cumulative mass of $p$, such that,
\begin{align*}
\sum_{j=1}^{t_i-1} R(\mathbf{x}_i)_j < p \le \sum_{j=1}^{t_i} R(\mathbf{x}_i)_j
\end{align*}.

Intuitively, $t_i$ is a function of the normalized scores --- it is defined exactly as in \citet{huang2024harderTopP}. While their work pretrained a model with a regularizer factor to ensure that $n_i$ is low on average, no such guarantee is assumed here. And thus, we decide to further cap $n_i$ by $k_0$. In general, it should never help to set $k_0 > k$ where $k$ is the model's default configuration. 
Finally, note that by setting $p=1$, we essentially have a fixed $k_0$ and by setting $k_0=N$, we practically have the top-$p$ method of \citet{huang2024harderTopP}, so we generalize and abstract on both methods. We decided to adopt this approach to allow the number of experts to be adaptive to the router scores so that harder instances can demand more experts.

This $(k_0, p)$-heuristic can select experts deemed critical to at least one token's predictions, and therefore the set of all essential experts $S^\text{base}=\cup_{i=1}^B S^{\text{base}}_i$ to activate.

\paragraph{Phase 2: Opportunistic piggybacking.} Instead of adding any new experts into the mix, this second phase opportunistically recovers some performance by allowing tokens to \emph{piggyback} onto experts already included in $S^\text{base}$, thus maintaining the number of activated experts $T=|S^\text{base}|$. For each token $i$, we traverse experts in decreasing order of their scores and select those in $S^\text{base}$ provided that (1) the number of selected experts does not exceed $k^\text{max}$ and (2) the expert's rank does not fall below a threshold position $\text{maxP}$. These constraints ensure that the selected experts do not degrade performance, either by over-diversifying expert usage or by selecting experts poorly aligned with the current token.

\paragraph{Weighting after rerouting.}
Once the routing sets $S_i$ are chosen, we keep the model’s original router scores and renormalize them following Equation~\eqref{eq:moeDefinition}.
Intuitively, this preserves the model’s learned preferences among the experts we keep, while ensuring mixture weights still sum to $1$. Other choices like using the weights of top-$k$ are possible, but we leave such optimization to future work.

\section{Experimental Setup And Empirical Results}
\paragraph{Hardware and model.} Unless otherwise specified, all our experiments were performed on one NVIDIA H100 80GB GPU each, while using Qwen3-30B-A3B \cite{yang2025qwen3} under \texttt{bfloat16} precision. This model has $48$ layers, each with $N=128$ experts of which $k=8$ are activated per token, $32$ query heads and $4$ KV heads, an embedding dimension of $2048$ and per-expert hidden dimension of $768$;
each expert uses SwiGLU-based \cite{shazeer2020SwiGLU} feedforward network which entails $3$ matrix multiplications of sizes $2048\times 768$.

\subsection{Cross-Entropy Experiments}

\label{sec:experiments-ppl}
\paragraph{Motivation.} We use cross-entropy on a pretraining dataset as a granular proxy for the compound effect of our router intervention. We do this because:
\begin{itemize}
    \item It is much cheaper to measure cross-entropy than downstream performance thanks to its parallel computation. Thus, we can perform a large hyperparameter sweep to determine the \emph{optimal} setting of \methodname. Based on these findings, we suggest a simplified version of \methodname~(Algorithm~\ref{alg:simplified-router}) that we then evaluate on standard benchmarks in Section~\ref{sec:downstream-eval}.
    \item Unlike benchmarks, cross-entropy provides a more statistically reliable estimate of modeling quality since it provides dense, per-token signal rather than sparse, task-level evaluation.
\end{itemize}

\paragraph{Dataset.} In the first round of experiments, we evaluated cross-entropy loss on a subset of the FineWeb-Edu dataset~\cite{penedo2024finewebEDU}, which we selected as a high-quality and diverse proxy since Qwen3's pretraining data is not public. We randomly selected $2048$ sequences, each containing at least $8192$ tokens to ensure a fixed batch size across positions since \methodname\ is sensitive by construction to batch size. In particular, a batch size of $1$ deems the piggybacking redundant.

\paragraph{Methodology.} We simulate $L$ decoding steps (sequence length) but execute them efficiently in parallel. At step $t$, we form a batch from the $t$-th token of each sequence and run routing \emph{only within that step}: both the Phase 1 pruning and Phase 2 piggybacking are computed using tokens that share the same position $t$.
No information (experts or scores) is shared across different positions, so piggybacking never crosses decode steps. We then process all steps in parallel by grouping expert workloads post-routing, which yields the same routing decisions as true sequential decode while enabling a fast, batched implementation for measurement. Throughout this computation, we track the \emph{average number of activated experts} across positions and layers, as well as \emph{average cross-entropy}.

\paragraph{Experiments.} The parallel speedup allows for a comprehensive sweep of hyperparameters: $k_0 \in \{4,5,6,7,8\}$, $k^\text{max} \in \{7,8,9,10,11\}$, $p \in \{0.4,0.5,0.6,0.7,0.8,0.9,1\}$ and $\text{maxP}\in \{8, 16, 32, 128\}$.
On top of these, we also considered stopping after Phase 1 (for the same ranges of $k_0, p$) and forgoing the piggybacking --- we refer to this as \emph{Phase 1} or \emph{pruned} routing. For each such routing algorithm, we swept $B \in \{8, 16, 32, 64\}$. We used $128$ sequences and the full length of $8192$ for $B=8$ and cut sequence length by half for every batch doubling to keep the activation memory fixed and able to fit in the memory of a GPU. We also doubled the number of sequences for every batch doubling to keep the overall number of tokens fixed at 1M.

Equipped with these runs, we can investigate the effects of our design choices. There are four degrees of freedom corresponding to the four hyperparameters: $k_0$ and $p$ determine the pruning extent while $k^\text{max}$ and $\text{maxP}$ control whether adding an extra expert starts hurting.\footnote{Note that setting $p=1$ is equivalent to not using it at all and so is the case for $\text{maxP}=128$.}

\paragraph{Ablations.}
For each experiment, we define its performance as a trade-off between cross-entropy and the average number of activated experts --- the objective is to minimize both.
To assess the effect of each hyperparameter value on performance, we plot the Pareto frontier of all experiments conducted with that value.
As our purpose is to limit cross-entropy degradation, we plot the increase in cross-entropy with respect to a vanilla MoE and track across runs. Furthermore, as a lot of runs present negligible differences in cross-entropy and average number of activated experts, we round the increase in cross-entropy to the closest multiple of $0.005$ and the average number of activated experts to the closest multiple of $0.1$ to avoid crowding the plots.

\begin{figure}[t]
  \centering
  \includegraphics[width=.9\linewidth]{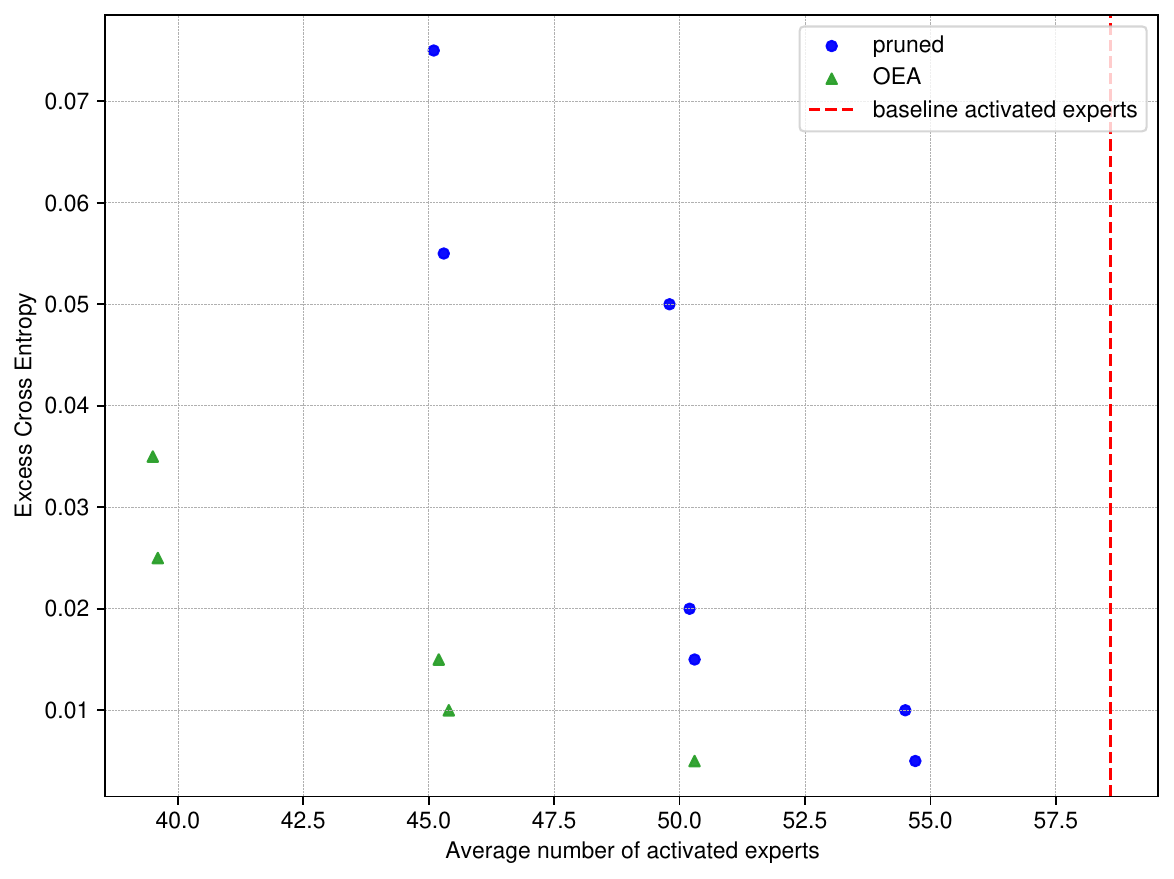}
  \caption{The y-axis shows the cross-entropy delta relative to the baseline (lower left is better). The two types of dots correspond to the Pareto frontiers of pruned and \methodname\ experiments at batch size $B=16$. \methodname\ consistently performs better.}
  \label{fig:pruned_vs_oae16_main}
\end{figure}

\paragraph{Piggybacking gains.} Since our algorithm's core addition over a form of adaptive pruning is the piggybacking phase, the salient question is whether Phase $2$ truly adds value: we answer this in the affirmative, as shown in Figure~\ref{fig:pruned_vs_oae16_main}.

We find three consistent patterns regarding hyperparameter choice. The ablation plots corresponding to them are available in the Appendix~\ref{append:more-results}.
\begin{enumerate}
\item \textbf{Using $p<1$ does not help.} Setting $p=1$ (equivalent to using top-$k_0$ in Phase $1$) performs on par with $p<1$ (Figure~\ref{fig:all-p}). This holds across both \methodname\ and the partial ``pruned'' (Phase $1$ only) case. We considered employing a top-$p$ scheme to allow the choice of base quality to be a function of the router scores, but there is no significant marginal gain from this adaptivity.

\item \textbf{$k^\text{max}=k$ works best.} Interestingly, we find that bounding the number of experts per token at exactly $k$ works better than both smaller and larger choices (Figure~\ref{fig:all-maxk}). Naturally, one expects more experts to help but interestingly, using $k^\text{max}=9$ experts does not really improve above $k^\text{max}=8$; in fact, further increasing to $k^\text{max}=10,11$ actually results in degradation. 

\item \textbf{Setting $maxP < N$ does not help.} Our ablation over $\text{maxP}$ (Figure~\ref{fig:all-maxp}) shows that refraining from piggybacking onto an activated expert due to it being too far down a token's preference list is detrimental for the optimal values of $k^\text{max}$.
It is worth noting that $\text{maxP}$ could only make a difference when we do not have $k^\text{max}$ activated experts in the top-$\text{maxP}$ preferences of a token, which becomes less likely with the increase in $B$ (and thus $|S^\text{base}|$).
Finally, one important consequence of $\text{maxP}=8$ strictly hurting is proving that using out-of-policy experts confers a strict advantage. This is contrary to the thesis that those experts are not trained to be useful for this one token.
\end{enumerate}

\begin{figure}[t]
  \centering
  \includegraphics[width=.9\linewidth]{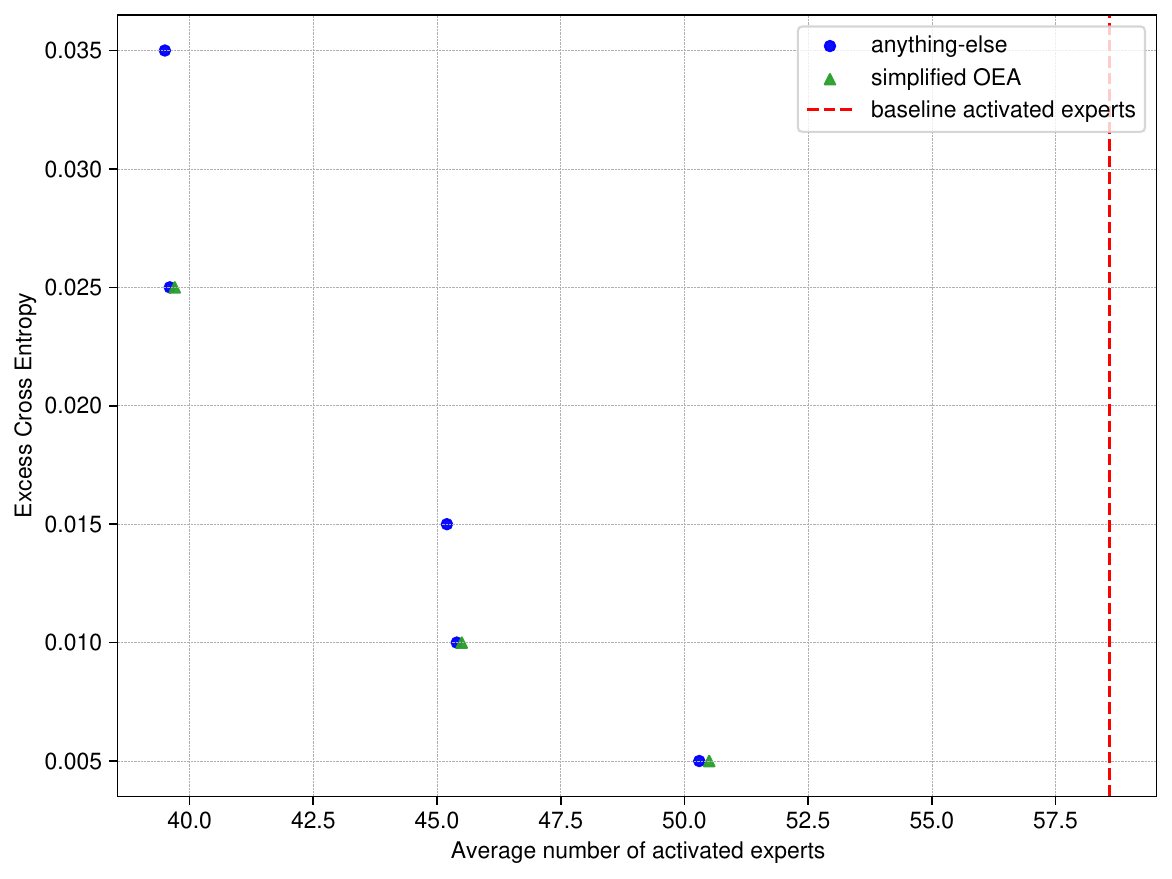}
  \caption{
  The y-axis shows the cross-entropy delta relative to the baseline (lower left is better). The two types of dots correspond to the Pareto frontiers of simplified OEA and the rest of experiments at batch size $B=16$. Simplified OEA performs comparably to the best hyperparameter choices.}
  \label{fig:simplified16_main}
\end{figure}
\paragraph{Simplifying \methodname} Putting these together, we conclude that we can drop the usage of top-$p$ in Phase $1$ and that of $\text{maxP}$ in Phase 2, as well as set $k^\text{max}$ to be $k$. This leaves us with a simplified version of the \methodname~routing which is presented in Algorithm~\ref{alg:simplified-router}. Figure~\ref{fig:simplified16_main} shows that our hyperparameter findings are jointly consistent, suggesting the simplified algorithm to be as performant as its general counterpart. A major benefit is therefore the reduced cost of hyperparameter sweeps prior to deploying the model: $k_0$ itself controls both the guaranteed baseline quality and the drop in activated experts.

\subsection{Downstream Evaluations}
\label{sec:downstream-eval}
\paragraph{Qwen3-235B-A22B model.} We benchmark our approach on both the Qwen3-30B and Qwen3-235B-A22B models. The latter doubles the number of layers ($96$), embedding dimension ($4096$) and expert hidden dimension ($1536$) while the same attention head configuration top-$8$/$128$ routing. All experiments are performed under tensor parallelism across $8$ H100 GPUs within a single HGX H100 node interconnected via NVSwitch ($18$ NVLink per GPU pair).

\paragraph{Setup.} We conducted downstream evaluation on four benchmarks: AIME 24, MATH 500~\citep{hendrycks2021measuringMATH500}, GPQA~\cite{rein2024gpqa} and LiveCodeBench~\cite{jain2024livecodebench}. All accuracy reported is an average over four runs of each for Qwen3-30B (and three runs for Qwen3-235B), with the exception of AIME 24 which we evaluate four times more runs (since it only has $30$ data points and, thus, higher variance). For all runs, we use a temperature of $0.6$, top-$p$ sampling with $p=0.95$ and generate up to $32768$ tokens. We integrate our router into the SGLang framework~\cite{zheng2024sglang}. For each run, we track the batch size, number of activated experts and the latency for every layer and decode step. Note that, throughout the serving process, batch size can and does vary as requests are finished, enqueued or retracted. Since our routing algorithm only benefits latency in moderate batch size regimes, we use it only during decode, not prefill. 

\paragraph{Experiments.} Informed by our findings from Section~\ref{sec:experiments-ppl}, we only used the recommended settings and thus only tried the simplified algorithm (\ref{alg:simplified-router}) parameterized only by $k_0$. We tested all values of $k_0 \in \{3,4,5,6,7\}$ on Qwen3-30B, and all except $k_0=7$ on Qwen3-235B due to computational constraints. We also evaluated post-Phase $1$ (pruned) routing (as in Section~\ref{sec:experiments-ppl}) for the same set of $k_0$ values.
As batch size cannot be fixed in SGLang, we use its {\small\texttt{--max-running-requests}} option to set a maximum batch size. We could not do this for a batch size of $32$ due to the large KV cache size, so we report \methodname\ numbers for a batch size bounded at $16$.

\paragraph{Piggybacking gains.} Tables~\ref{tab:30b-accuracies-merged-bold} and~\ref{tab:235b-accuracies-merged-bold} report the average performances of the pruned (Phase $1$) approach, of the simplified \methodname, and of the base model ($k_0$=8) for the 30B and 235B model, respectively. We find that:
\begin{itemize}
    \item \textbf{Qwen3-30B}: Across all benchmarks, except GPQA, using top-$5$ rather than top-$8$ experts does not make a statistically significant difference\footnote{A result $\mu \pm \se$ is considered standard-error adjusted worse than $\mu_\text{vanilla} \pm \se_\text{vanilla}$ when $\mu + \se < \mu_\text{vanilla} - \se_\text{vanilla}$}. However, further lowering it to $4$ and $3$ starts showing substantial degradation. \methodname\ manages to recover lost performance even for $k_0=3$ while fundamentally not incurring any extra cost over its pruned counterpart; this is the marginal gain of piggybacking and our main contribution.
    \item \textbf{Qwen3-235B}: In the basic pruning setup, performance drops sharply at $k_0=5$ falling below the base model by $15$\% on LiveCodeBench, $8$\% on GPQA. In contrast, \methodname\ at $k_0=5$ maintains performance on all benchmarks except for LiveCodeBench where its accuracy declines slightly by $2$\%.
\end{itemize}

\paragraph{Relationship between latency and the number of activated experts.} The central hypothesis of this work (introduced in Section~\ref{sec:goal-rephrasing}) is that for moderate batch sizes, the MoE latency scales linearly with the number of activated experts. To confirm this, we tracked all the ($T$, latency) pairs obtained at all decode steps and all layers.
Figure~\ref{fig:expert_latency_gpqa} shows the average latency for a fixed number of activated experts across the whole GPQA run of the vanilla Qwen3-30B model (across decode steps and layers); the standard errors are all less than $2\cdot10^{-4}$ indicating that latency is well predicted by these means. Since the MoE module itself was left unchanged, this trend is independent of the routing strategy. Strikingly, the linear trend fits regression at $R^2 > 0.99$, thus affirming the thesis of this work: latency is linearly controlled by the number of activated experts.

\paragraph{Latency gains via reducing active experts.}
Building on the above, we examine how \methodname’s reduced expert activation translates to practical latency reductions.
For the Qwen3-30B model, Table~\ref{table:oae30b-avg-experts} reports the average number of active experts (aggregated over layers and decode steps) as a function of $k_0$, with $k_0=3$ halving the number of activated experts. As shown in Table~\ref{table:oae30b-latency}, this corresponds to latency reductions of $39$\% for $k_0=3$ and $23$\% for $k_0=5$.
For Qwen3-235B, the results in Table~\ref{table:oae235b-latency} show a $15$\% speedup at $k_0=5$; we attribute this smaller relative reduction to the additional overhead of tensor parallel's all-reduce.

\begin{table*}[t]
\caption{Average MoE layer latency (in microseconds) when using \textbf{simplified OEA} (top-$k_0$ + piggybacking) on Qwen3-30B-A3B.}
\label{table:oae30b-latency}
\vskip 0.15in
\begin{center}
\begin{small}
\begin{sc}
\begin{tabular}{lcccccc}
\toprule
 & $k_0=3$ & $k_0=4$ & $k_0=5$ & $k_0=6$ & $k_0=7$ & vanilla \\
\midrule
aime24 & 97.9 & 110.5 & 122.0 & 138.4 & 147.4 & 158.0 \\
gpqa & 111.0 & 125.4 & 143.2 & 159.0 & 172.5 & 184.1 \\
livecodebench & 102.7 & 117.1 & 132.2 & 146.5 & 157.3 & 170.8 \\
math\_500 & 115.6 & 130.4 & 146.7 & 161.5 & 174.7 & 189.9 \\
\midrule
average & 106.8 & 120.9 & 136.0 & 151.3 & 163.0 & 175.7 \\
normalized average & 0.61 & 0.69 & 0.77 & 0.86 & 0.93 & 1.00 \\
\bottomrule
\end{tabular}
\end{sc}
\end{small}
\end{center}
\vskip -0.1in
\end{table*}

\begin{table*}[t]
\caption{Average number of activated experts when using \textbf{simplified OEA} (top-$k_0$ + piggybacking) on Qwen3-30B-A3B.}
\label{table:oae30b-avg-experts}
\vskip 0.15in
\begin{center}
\begin{small}
\begin{sc}
\begin{tabular}{lcccccc}
\toprule
 & $k_0=3$ & $k_0=4$ & $k_0=5$ & $k_0=6$ & $k_0=7$ & vanilla \\
\midrule
aime24 & 22.2 & 26.5 & 30.5 & 36.0 & 39.2 & 43.0 \\
gpqa & 26.5 & 31.4 & 37.5 & 42.9 & 47.6 & 51.6 \\
livecodebench & 23.8 & 28.7 & 33.9 & 38.7 & 42.5 & 47.2 \\
math\_500 & 27.9 & 33.0 & 38.6 & 43.7 & 48.3 & 53.5 \\
\midrule
average & 25.1 & 29.9 & 35.1 & 40.3 & 44.4 & 48.8 \\
normalized average & 0.51 & 0.61 & 0.72 & 0.83 & 0.91 & 1.00 \\
\bottomrule
\end{tabular}
\end{sc}
\end{small}
\end{center}
\vskip -0.1in
\end{table*}
\begin{table*}[t]
\caption{Average MoE layer latency (in microseconds), including \emph{all-reduce} when using simplified OEA (top-$k_0$ + piggybacking) on Qwen3-235B-A22B.}
\label{table:oae235b-latency}
\vskip 0.15in
\begin{center}
\begin{small}
\begin{sc}
\begin{tabular}{lccccc}
\toprule
 & $k_0=3$ & $k_0=4$ & $k_0=5$ & $k_0=6$ & vanilla \\
\midrule
aime24 & 86.4 & 92.6 & 98.8 & 105.7 & 118.4 \\
gpqa & 86.7 & 93.8 & 99.5 & 104.7 & 116.0 \\
livecodebench & 88.2 & 95.3 & 102.8 & 108.6 & 121.1 \\
math\_500 & 89.6 & 97.4 & 104.4 & 108.7 & 122.2 \\
\midrule
average & 87.7 & 94.8 & 101.4 & 106.9 & 119.4 \\
normalized average & 0.73 & 0.79 & 0.85 & 0.90 & 1.00 \\
\bottomrule
\end{tabular}
\end{sc}
\end{small}
\end{center}
\vskip -0.1in
\end{table*}

\section{Related Work}
\label{sec:related-work}
The challenge of optimizing MoE inference latency is an active area of research \cite{liu2024surveyMoEInference}. \methodname\ builds upon lessons from works on alternative routing mechanisms, architectural innovations, and other dynamic inference-time strategies, while distinguishing in several ways.

\subsection{Foundational MoE Routing and Its System-Level Challenges}

The modern MoE paradigm in Transformers was established by Shazeer et al. with the Sparsely-Gated MoE layer \yrcite{shazeer2017outrageouslyFirstMoE}, which employs a trainable gating network to route each token to a top-$k$ subset of experts. Their goal was to decouple the model parameters from the computational cost of training to enable scaling of larger LLMs.

However, this approach introduces significant systems-level challenges. The most notable is \textbf{load imbalance} (where the router ends up favoring a subset of ``popular" experts) leads to router collapse, leaving other experts and their associated parameters and hardware underutilized. To address this matter, load-balancing losses are employed during training \cite{shazeer2017outrageouslyFirstMoE,fedus2022switch,liu2024deepseek}.

More central to our work are the issues that arise in batched inference for large sparsity. Serving systems \cite{kwon2023efficientVLLM,zheng2024sglang} rely on batching to achieve high throughput, but this forces the activation of the union of all experts selected by any token in the batch, quickly negating MoE's sparsity and deeming MoE layers to be memory-bound for moderate batch sizes.

\subsection{Alternative Routing and Architectural Paradigms}

To address these fundamental issues, paradigms going beyond token-centric top-$k$ routing have been explored.

\paragraph{Expert choice routing.} \citet{zhou2022mixtureExpertChoiceRouting} inverted the selection logic, allowing each expert to select its top-$k$ preferred tokens from the batch. This is an inherently batch-aware mechanism that guarantees perfect load balancing by design, eliminating the need for auxiliary losses. While it enables a variable number of experts per token, its purpose is optimizing throughput via load balancing rather than minimizing the number of active experts to reduce latency.

\paragraph{Architectural solutions (shared experts).} Models like DeepSeek-V3~\cite{liu2024deepseek} and Kimi K2~\cite{team2025kimi} incorporate a hybrid architecture with both ``routed'' and ``shared'' experts. The shared experts process every token in the batch, providing a form of guaranteed computational reused core. This allows for system co-optimization, such as hiding communication latency behind the shared expert's computation. Such approaches represent a static design solution to shared computation, contrasting with our dynamic, runtime approach that requires no architectural modifications.

\subsection{Dynamic Inference-Time Optimizations}

\methodname\ best fits under the paradigm of dynamic, inference-time optimizations that modify MoE behavior without retraining. Such approaches are crucially different from static pruning methods that permanently remove experts they expect to not be crucial to performance. While effective for compression, the behavior of the dropped experts cannot be recovered if a token depended on them. Such issues could in principle be mitigated by making exclusion decisions adaptive to the batch, although then the model size cannot be reduced.

\paragraph{Token-centric dynamic skipping.} One category of dynamic methods operates on a per-token basis. Lu et al.~\yrcite{lu2024notAllExpertsEq} proposed dynamically skipping a secondary expert if its router score is significantly lower than the primary one, saving computation on a per-token basis. Similarly, the Online Dynamic Pruning (ODP) technique identifies less important tokens and assigns them fewer experts \cite{huang2024mixtureCompressorOnlineDynamicPruning}. These methods are not explicitly batch-aware and thus miss opportunities for shared computation.

\paragraph{Expert offloading and prefetching.} In environments that are memory constrained environments, model weights are stored on CPU and offloaded on a need basis. Systems like Pre-gated MoE~\cite{hwang2024preGatedMoe} employ predictive prefetching, where information from the current layer is used to anticipate and pre-load experts for the next layer while the current layer's computation is underway. These systems optimize for memory transfer costs across time (inter-batch), whereas our method optimizes for computational reuse within a single batch.

\paragraph{Comparison to related work.} The most closely related work is Lynx, a framework that also uses a batch-aware routing to reduce active experts \cite{gupta2024lynx}. They employ a fundamentally different, subtractive approach: first the union of experts that would normally be activated is computed; then, the least popular among these experts are dropped. Crucially, this risks removing an expert that, while unpopular across the batch, is critical to a single token's accuracy. In contrast, \methodname\ is an additive and opportunistic framework. It first guarantees the critical computation for every token, ensuring a baseline quality. It then opportunistically augments this baseline by ``piggybacking'' on experts that are already active, recovering model capacity at \emph{zero} additional latency cost. This additive approach provides a more robust trade-off between performance and accuracy, particularly for modern models with a high number of activated experts ($k > 2$), where the hierarchical importance of experts observed by Lynx may be less pronounced. Finally, note that our piggybacking phase can be added to any routing-changing approach (static or dynamic) --- including Lynx --- to gain \emph{free} quality recovery.

\section{Discussion}
\paragraph{Effect of batch distribution.} \methodname's effectiveness depends strongly on the tokens' distribution within a batch. When tokens come from similar distributions, they tend to overlapping experts resulting in a smaller $S^\text{base}$ which limits piggybacking's gains. This is the regime that our benchmarks represent, making the reported performance a conservative estimate. In contrast, the cross-entropy experiments in Section~\ref{sec:experiments-ppl} correspond to a more diverse token distribution which enlarges $S^\text{base}$ and allows piggybacking to recover more of the base model's performance.

\paragraph{A note on padding.} During our experiments, under default configuration of SGLang, we noticed the average number of tokens (and average latency) in batches of size $7$ exceed that of batches of size $8$, which was counter-intuitive. This is because SGLang captures CUDA Graphs for a set of batch-sizes and when it needs to process a certain batch size $B$, it looks up the smallest $B'>B$ that has been captured and pads the batch up to size $B'$. While for classic feed-forward networks and attention, the specific contents of the batch do not influence the kernel's runtime, this is not the case for MoEs, especially under the memory-bound batch regime we operate. What happened was that the padding token activated on average more experts ``out-of-distribution'' (that were not activated by real tokens already) than an $8$\textsuperscript{th} realistic one would. Thus this seemingly inoffensive padding ended up costing more than processing an extra real token, when it should ideally not add any more experts. In our experiments we simply fixed this by capturing CUDA Graphs up to size $16$ (thus ensuring no padding), but we do make the general note that there is value in adding a padding mask and using it to zero out the padding tokens' expert choices.

\section{Conclusion and Future Directions}
In this work, we introduce \methodname, a new expert routing algorithm targeting the problem of memory-bound MoE decoding under moderate batch sizes. \methodname~achieves this by reducing the number of activated experts per decode batch. To do this, it first activates a few top experts per token deemed crucial to its performance and then opportunistically piggybacks some more that were crucial to other tokens in the batch. This approach results in MoE latency speed-ups of $39$\% and $15$\% for the Qwen3-30B and Qwen3-235B models, respectively, without statistically significant degradation in benchmark accuracy.

\paragraph{Batch adaptivity.} The cross-entropy's analysis (Figure~\ref{fig:all-pruned-vs-oae}d) shows negligible degradation at $B=64$. Larger batches naturally increase $S^\text{base}$, enabling piggybacking to approximate the original routing more closely.
This observation suggests an approach where the routing scheme is a function of the batch-size (e.g. using a bigger (safer) $k_0$ at a lower batch size). We leave determining such batch-size-dependent $k_0$-choice as an open problem.

\paragraph{Extension to expert parallelism.} Although \methodname\ assumes experts are not executed in parallel, adapting to the expert-parallel setting is possible --- the latency is then driven by the \emph{maximum} number of activated experts per machine. Hence, one immediate equivalent to our method is to do piggybacking independently on each machine while potentially increasing $k_0$ in the machines that activate fewer experts (have a smaller $S^\text{base}$).

\paragraph{Layer heterogeneity.} Empirically, we observe that the average number of active experts varies significantly across layers. This, coupled with previous empirical findings~\cite{gupta2024lynx,yang2025fasterStaticLayerHeteroPruning}, suggests that adapting our method's hyperparameters independently for each layer could further improve its performance.

\paragraph{Routing robustness and model co-designing.} At its core, our method relies on pretrained models' routing being robust to slight changes. Therefore, further understanding or quantifying the routing robustness limits, or even having models' pretraining be co-designed with this purpose in mind could further increase router flexibility and thus improve \methodname's scope. This includes tackling the challenge mentioned in Section~\ref{sec:algorithm} of how to adapt expert weights when expert routing is modified.

\bibliography{real_paper}

\begin{thebibliography}{25}
\providecommand{\natexlab}[1]{#1}
\providecommand{\url}[1]{\texttt{#1}}
\expandafter\ifx\csname urlstyle\endcsname\relax
  \providecommand{\doi}[1]{doi: #1}\else
  \providecommand{\doi}{doi: \begingroup \urlstyle{rm}\Url}\fi

\bibitem[Fedus et~al.(2022)Fedus, Zoph, and Shazeer]{fedus2022switch}
Fedus, W., Zoph, B., and Shazeer, N.
\newblock Switch transformers: Scaling to trillion parameter models with simple and efficient sparsity.
\newblock \emph{Journal of Machine Learning Research}, 23\penalty0 (120):\penalty0 1--39, 2022.

\bibitem[Gupta et~al.(2024)Gupta, Sinha, Gavrilovska, and Iyer]{gupta2024lynx}
Gupta, V., Sinha, K., Gavrilovska, A., and Iyer, A.~P.
\newblock Lynx: Enabling efficient moe inference through dynamic batch-aware expert selection.
\newblock \emph{arXiv preprint arXiv:2411.08982}, 2024.

\bibitem[Hejazi(2024)]{Hejazi2024_GroupedGEMM}
Hejazi, B.
\newblock Introducing grouped gemm apis in cublas and more performance updates.
\newblock \url{https://developer.nvidia.com/blog/introducing-grouped-gemm-apis-in-cublas-and-more-performance-updates/}, June 2024.
\newblock NVIDIA Developer Blog, June 12, 2024.

\bibitem[Hendrycks et~al.(2021)Hendrycks, Burns, Kadavath, Arora, Basart, Tang, Song, and Steinhardt]{hendrycks2021measuringMATH500}
Hendrycks, D., Burns, C., Kadavath, S., Arora, A., Basart, S., Tang, E., Song, D., and Steinhardt, J.
\newblock Measuring mathematical problem solving with the math dataset.
\newblock \emph{arXiv preprint arXiv:2103.03874}, 2021.

\bibitem[Huang et~al.(2024{\natexlab{a}})Huang, An, Zhuang, Tao, Zhang, Jin, Xu, Chen, Huang, and Feng]{huang2024harderTopP}
Huang, Q., An, Z., Zhuang, N., Tao, M., Zhang, C., Jin, Y., Xu, K., Chen, L., Huang, S., and Feng, Y.
\newblock Harder tasks need more experts: Dynamic routing in moe models.
\newblock \emph{arXiv preprint arXiv:2403.07652}, 2024{\natexlab{a}}.

\bibitem[Huang et~al.(2024{\natexlab{b}})Huang, Liao, Liu, He, Tan, Zhang, Li, Liu, and Qi]{huang2024mixtureCompressorOnlineDynamicPruning}
Huang, W., Liao, Y., Liu, J., He, R., Tan, H., Zhang, S., Li, H., Liu, S., and Qi, X.
\newblock Mixture compressor for mixture-of-experts llms gains more.
\newblock \emph{arXiv preprint arXiv:2410.06270}, 2024{\natexlab{b}}.

\bibitem[Hwang et~al.(2024)Hwang, Wei, Cao, Hwang, Tang, Cao, and Yang]{hwang2024preGatedMoe}
Hwang, R., Wei, J., Cao, S., Hwang, C., Tang, X., Cao, T., and Yang, M.
\newblock Pre-gated moe: An algorithm-system co-design for fast and scalable mixture-of-expert inference.
\newblock In \emph{2024 ACM/IEEE 51st Annual International Symposium on Computer Architecture (ISCA)}, pp.\  1018--1031. IEEE, 2024.

\bibitem[Jain et~al.(2024)Jain, Han, Gu, Li, Yan, Zhang, Wang, Solar-Lezama, Sen, and Stoica]{jain2024livecodebench}
Jain, N., Han, K., Gu, A., Li, W.-D., Yan, F., Zhang, T., Wang, S., Solar-Lezama, A., Sen, K., and Stoica, I.
\newblock Livecodebench: Holistic and contamination free evaluation of large language models for code.
\newblock \emph{arXiv preprint arXiv:2403.07974}, 2024.

\bibitem[{Kimi Team} et~al.(2025){Kimi Team}, Bai, Bao, Chen, Chen, Chen, Chen, Chen, Chen, Chen, et~al.]{team2025kimi}
{Kimi Team}, Bai, Y., Bao, Y., Chen, G., Chen, J., Chen, N., Chen, R., Chen, Y., Chen, Y., Chen, Y., et~al.
\newblock {Kimi K2: O}pen agentic intelligence.
\newblock \emph{arXiv preprint arXiv:2507.20534}, 2025.

\bibitem[Kwon et~al.(2023)Kwon, Li, Zhuang, Sheng, Zheng, Yu, Gonzalez, Zhang, and Stoica]{kwon2023efficientVLLM}
Kwon, W., Li, Z., Zhuang, S., Sheng, Y., Zheng, L., Yu, C.~H., Gonzalez, J., Zhang, H., and Stoica, I.
\newblock Efficient memory management for large language model serving with pagedattention.
\newblock In \emph{Proceedings of the 29th symposium on operating systems principles}, pp.\  611--626, 2023.

\bibitem[Li et~al.(2025)Li, Li, and Zhou]{li2025r2multimodalReroute}
Li, Z., Li, Z., and Zhou, T.
\newblock R2-t2: Re-routing in test-time for multimodal mixture-of-experts.
\newblock \emph{arXiv preprint arXiv:2502.20395}, 2025.

\bibitem[Liu et~al.(2024{\natexlab{a}})Liu, Feng, Xue, Wang, Wu, Lu, Zhao, Deng, Zhang, Ruan, et~al.]{liu2024deepseek}
Liu, A., Feng, B., Xue, B., Wang, B., Wu, B., Lu, C., Zhao, C., Deng, C., Zhang, C., Ruan, C., et~al.
\newblock Deepseek-v3 technical report.
\newblock \emph{arXiv preprint arXiv:2412.19437}, 2024{\natexlab{a}}.

\bibitem[Liu et~al.(2024{\natexlab{b}})Liu, Zhu, Lin, Ning, Blaschko, Yan, Dai, Yang, and Wang]{liu2024evolutionaryStaticPruning}
Liu, E., Zhu, J., Lin, Z., Ning, X., Blaschko, M.~B., Yan, S., Dai, G., Yang, H., and Wang, Y.
\newblock Efficient expert pruning for sparse mixture-of-experts language models: Enhancing performance and reducing inference costs.
\newblock \emph{arXiv preprint arXiv:2407.00945}, 2024{\natexlab{b}}.

\bibitem[Liu et~al.(2024{\natexlab{c}})Liu, Tang, Wang, Ren, Hou, Heng, Guo, and Li]{liu2024surveyMoEInference}
Liu, J., Tang, P., Wang, W., Ren, Y., Hou, X., Heng, P.-A., Guo, M., and Li, C.
\newblock A survey on inference optimization techniques for mixture of experts models.
\newblock \emph{arXiv preprint arXiv:2412.14219}, 2024{\natexlab{c}}.

\bibitem[Lu et~al.(2024)Lu, Liu, Xu, Zhou, Huang, Zhang, Yan, and Li]{lu2024notAllExpertsEq}
Lu, X., Liu, Q., Xu, Y., Zhou, A., Huang, S., Zhang, B., Yan, J., and Li, H.
\newblock Not all experts are equal: Efficient expert pruning and skipping for mixture-of-experts large language models.
\newblock \emph{arXiv preprint arXiv:2402.14800}, 2024.

\bibitem[NVIDIA(2022)]{nvidia2022h100WhitePaper}
NVIDIA.
\newblock H100 tensor core gpu architecture white paper.
\newblock Technical report, NVIDIA Corporation, 2022.
\newblock URL \url{https://resources.nvidia.com/en-us-hopper-architecture/nvidia-h100-tensor-c}.

\bibitem[Penedo et~al.(2024)Penedo, Kydl{\'\i}{\v{c}}ek, Lozhkov, Mitchell, Raffel, Von~Werra, Wolf, et~al.]{penedo2024finewebEDU}
Penedo, G., Kydl{\'\i}{\v{c}}ek, H., Lozhkov, A., Mitchell, M., Raffel, C.~A., Von~Werra, L., Wolf, T., et~al.
\newblock The fineweb datasets: Decanting the web for the finest text data at scale.
\newblock \emph{Advances in Neural Information Processing Systems}, 37:\penalty0 30811--30849, 2024.

\bibitem[Rajbhandari et~al.(2022)Rajbhandari, Li, Yao, Zhang, Aminabadi, Awan, Rasley, and He]{rajbhandari2022deepspeedMoE}
Rajbhandari, S., Li, C., Yao, Z., Zhang, M., Aminabadi, R.~Y., Awan, A.~A., Rasley, J., and He, Y.
\newblock Deepspeed-moe: Advancing mixture-of-experts inference and training to power next-generation ai scale.
\newblock In \emph{International conference on machine learning}, pp.\  18332--18346. PMLR, 2022.

\bibitem[Rein et~al.(2024)Rein, Hou, Stickland, Petty, Pang, Dirani, Michael, and Bowman]{rein2024gpqa}
Rein, D., Hou, B.~L., Stickland, A.~C., Petty, J., Pang, R.~Y., Dirani, J., Michael, J., and Bowman, S.~R.
\newblock Gpqa: A graduate-level google-proof q\&a benchmark.
\newblock In \emph{First Conference on Language Modeling}, 2024.

\bibitem[Shazeer(2020)]{shazeer2020SwiGLU}
Shazeer, N.
\newblock Glu variants improve transformer.
\newblock \emph{arXiv preprint arXiv:2002.05202}, 2020.

\bibitem[Shazeer et~al.(2017)Shazeer, Mirhoseini, Maziarz, Davis, Le, Hinton, and Dean]{shazeer2017outrageouslyFirstMoE}
Shazeer, N., Mirhoseini, A., Maziarz, K., Davis, A., Le, Q., Hinton, G., and Dean, J.
\newblock Outrageously large neural networks: The sparsely-gated mixture-of-experts layer.
\newblock \emph{arXiv preprint arXiv:1701.06538}, 2017.

\bibitem[Yang et~al.(2025{\natexlab{a}})Yang, Li, Yang, Zhang, Hui, Zheng, Yu, Gao, Huang, Lv, et~al.]{yang2025qwen3}
Yang, A., Li, A., Yang, B., Zhang, B., Hui, B., Zheng, B., Yu, B., Gao, C., Huang, C., Lv, C., et~al.
\newblock Qwen3 technical report.
\newblock \emph{arXiv preprint arXiv:2505.09388}, 2025{\natexlab{a}}.

\bibitem[Yang et~al.(2025{\natexlab{b}})Yang, Shi, Li, Li, Wang, Du, Shen, and Zhao]{yang2025fasterStaticLayerHeteroPruning}
Yang, H., Shi, L., Li, Q., Li, Z., Wang, P., Du, B., Shen, M., and Zhao, H.
\newblock Faster moe llm inference for extremely large models.
\newblock \emph{arXiv preprint arXiv:2505.03531}, 2025{\natexlab{b}}.

\bibitem[Zheng et~al.(2024)Zheng, Yin, Xie, Sun, Huang, Yu, Cao, Kozyrakis, Stoica, Gonzalez, et~al.]{zheng2024sglang}
Zheng, L., Yin, L., Xie, Z., Sun, C.~L., Huang, J., Yu, C.~H., Cao, S., Kozyrakis, C., Stoica, I., Gonzalez, J.~E., et~al.
\newblock Sglang: Efficient execution of structured language model programs.
\newblock \emph{Advances in neural information processing systems}, 37:\penalty0 62557--62583, 2024.

\bibitem[Zhou et~al.(2022)Zhou, Lei, Liu, Du, Huang, Zhao, Dai, Le, Laudon, et~al.]{zhou2022mixtureExpertChoiceRouting}
Zhou, Y., Lei, T., Liu, H., Du, N., Huang, Y., Zhao, V., Dai, A.~M., Le, Q.~V., Laudon, J., et~al.
\newblock Mixture-of-experts with expert choice routing.
\newblock \emph{Advances in Neural Information Processing Systems}, 35:\penalty0 7103--7114, 2022.

\end{thebibliography}
\bibliographystyle{mlsys2025}
\appendix
\newcommand{\sweepquad}[3]{%
  \begin{figure*}[t]
    \centering
    \subfigure[$B=8$]{%
      \includegraphics[width=0.48\linewidth]{sweep_plots/#1/bs8.pdf}%
    }\hfill
    \subfigure[$B=16$]{%
      \includegraphics[width=0.48\linewidth]{sweep_plots/#1/bs16.pdf}%
    }\\[0.5em]
    \subfigure[$B=32$]{%
      \includegraphics[width=0.48\linewidth]{sweep_plots/#1/bs32.pdf}%
    }\hfill
    \subfigure[$B=64$]{%
      \includegraphics[width=0.48\linewidth]{sweep_plots/#1/bs64.pdf}%
    }
    \caption{#2}
    \label{#3}
  \end{figure*}%
}

\section{More benchmark results}
\label{append:more-results}
Tables~\ref{tab:30b-oae-std} and~\ref{tab:235b-oae-std} show the benchmark accuracies of \emph{simplified OEA} together \emph{with the standard errors} across the $4$ Qwen3-30B, respectively $3$ Qwen3-235B runs.

Tables~\ref{tab:30b-pruned-std} and~\ref{tab:235b-pruned-std} show the benchmark accuracies for the \emph{pruned} routers (top-$k_0$) \emph{with the standard errors} across the $4$ Qwen3-30B, respectively, $3$ Qwen3-235B runs.

Furthermore, Table~\ref{table:oae235b-avg-experts} shows the average number of activated experts while using simplified OEA routing on Qwen3-235B.

\begin{figure}[b]
  \centering
  \includegraphics[width=.9\linewidth]{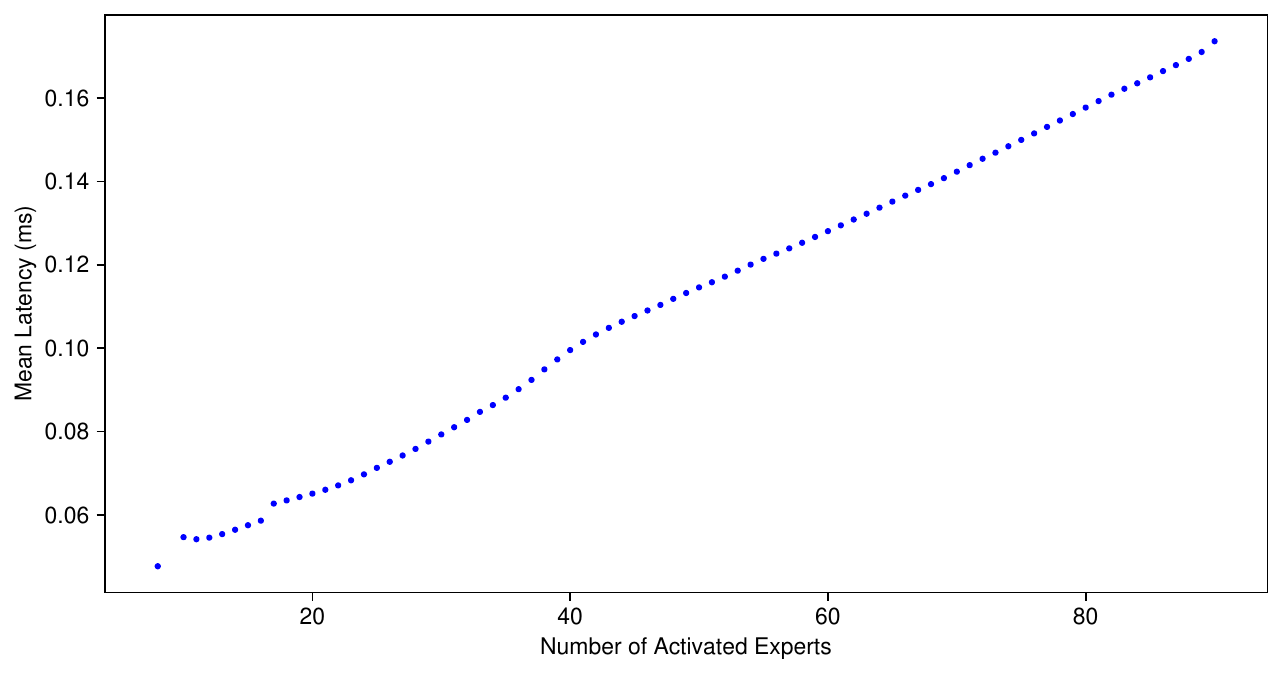}
  \caption{Mean MoE latency as a function of the number of activated experts within a decode batch. The average is computed over all layers and decode steps across a GPQA evaluation of the vanilla Qwen3-235B-A22B model (under a tensor parallel degree of $8$).}
  \label{fig:expert_latency_gpqa_235B}
\end{figure}

For the Qwen3-235B model, we report, in Figure~\ref{fig:expert_latency_gpqa_235B}, the average latency (across decode steps and layers) for a fixed number of activated experts across a complete run of GPQA.

\begin{table*}[t]
\caption{Ablation across $k_0$: Benchmark accuracies, \emph{standard error included}, of \textbf{simplified OEA} routing (top-$k_0$+piggybacking) on Qwen3-30B-A3B. Vanilla represents the default model.}
\label{tab:30b-oae-std}
\vskip 0.15in
\begin{center}
\begin{small}
\begin{sc}
\begin{tabular}{lcccccc}
\toprule
 & $k_0=3$ & $k_0=4$ & $k_0=5$ & $k_0=6$ & $k_0=7$ & vanilla \\
\midrule
aime24 & $80.0\pm 0.59$ & $81.9\pm 0.71$ & $81.5\pm 0.62$ & $80.8\pm 0.90$ & $78.5\pm 1.33$ & $80.4\pm 0.99$ \\
gpqa & $58.6\pm 1.22$ & $59.3\pm 0.15$ & $61.1\pm 1.69$ & $62.2\pm 0.56$ & $60.6\pm 0.36$ & $60.2\pm 0.83$ \\
livecodebench & $61.2\pm 0.94$ & $62.7\pm 0.53$ & $62.0\pm 0.33$ & $63.1\pm 0.57$ & $62.5\pm 0.47$ & $62.1\pm 0.94$ \\
math\_500 & $93.5\pm 0.29$ & $93.1\pm 0.29$ & $93.3\pm 0.27$ & $93.1\pm 0.30$ & $93.2\pm 0.08$ & $92.8\pm 0.22$ \\
\bottomrule
\end{tabular}
\end{sc}
\end{small}
\end{center}
\vskip -0.1in
\end{table*}

\begin{table*}[t]
\caption{Ablation across $k_0$: Benchmark accuracies, \emph{standard error included}, of \textbf{pruned} routing (top-$k_0$) on Qwen3-30B-A3B. Vanilla represents the default model.}
\label{tab:30b-pruned-std}
\vskip 0.15in
\begin{center}
\begin{small}
\begin{sc}
\begin{tabular}{lcccccc}
\toprule
 & $k_0=3$ & $k_0=4$ & $k_0=5$ & $k_0=6$ & $k_0=7$ & vanilla \\
\midrule
aime24 & $51.2\pm 1.42$ & $75.8\pm 0.83$ & $80.6\pm 0.86$ & $80.2\pm 0.40$ & $82.5\pm 0.83$ & $80.4\pm 0.99$ \\
gpqa & $45.7\pm 0.84$ & $54.3\pm 0.53$ & $56.2\pm 0.43$ & $58.3\pm 0.25$ & $59.7\pm 2.01$ & $60.2\pm 0.83$ \\
livecodebench & $37.4\pm 0.83$ & $58.2\pm 1.41$ & $63.2\pm 0.95$ & $63.1\pm 0.76$ & $63.0\pm 0.77$ & $62.1\pm 0.94$ \\
math\_500 & $91.1\pm 0.37$ & $92.7\pm 0.26$ & $92.6\pm 0.13$ & $93.1\pm 0.29$ & $93.3\pm 0.26$ & $92.8\pm 0.22$ \\
\bottomrule
\end{tabular}
\end{sc}
\end{small}
\end{center}
\vskip -0.1in
\end{table*}

\begin{table*}[t]
\caption{Ablation across $k_0$: Benchmark accuracies, \emph{standard error included}, of \textbf{simplified OEA} routing (top-$k_0$+piggybacking) on Qwen3-235B-A22B. Vanilla represents the default model.}
\label{tab:235b-oae-std}
\vskip 0.15in
\begin{center}
\begin{small}
\begin{sc}
\begin{tabular}{lccccc}
\toprule
 & $k_0=3$ & $k_0=4$ & $k_0=5$ & $k_0=6$ & vanilla \\
\midrule
aime24 & $81.4\pm 1.21$ & $82.5\pm 0.48$ & $83.6\pm 1.39$ & $83.6\pm 1.11$ & $85.0\pm 0.96$ \\
gpqa & $66.3\pm 0.67$ & $67.7\pm 1.05$ & $67.5\pm 0.67$ & $67.5\pm 0.45$ & $68.4\pm 0.34$ \\
livecodebench & $63.4\pm 0.75$ & $67.1\pm 0.73$ & $66.1\pm 0.63$ & $66.1\pm 0.12$ & $68.5\pm 0.21$ \\
math\_500 & $94.4\pm 0.46$ & $94.8\pm 0.31$ & $94.7\pm 0.13$ & $94.3\pm 0.18$ & $94.7\pm 0.18$ \\
\bottomrule
\end{tabular}
\end{sc}
\end{small}
\end{center}
\vskip -0.1in
\end{table*}
\begin{table*}[t]
\caption{Ablation across $k_0$: Benchmark accuracies, \emph{standard error included}, of \textbf{pruned} routing (top-$k_0$) on Qwen3-235B-A22B. Vanilla represents the default model.}
\label{tab:235b-pruned-std}
\vskip 0.15in
\begin{center}
\begin{small}
\begin{sc}
\begin{tabular}{lccccc}
\toprule
 & $k_0=3$ & $k_0=4$ & $k_0=5$ & $k_0=6$ & vanilla \\
\midrule
aime24 & $17.5\pm 0.96$ & $69.4\pm 0.73$ & $81.9\pm 1.39$ & $82.8\pm 1.21$ & $85.0\pm 0.96$ \\
gpqa & $43.8\pm 0.61$ & $56.4\pm 0.73$ & $60.6\pm 1.54$ & $64.1\pm 1.01$ & $68.4\pm 0.34$ \\
livecodebench & $5.7\pm 0.72$ & $27.4\pm 1.19$ & $53.5\pm 0.73$ & $60.8\pm 0.24$ & $68.5\pm 0.21$ \\
math\_500 & $80.9\pm 0.18$ & $93.3\pm 0.27$ & $94.5\pm 0.44$ & $94.5\pm 0.18$ & $94.7\pm 0.18$ \\
\bottomrule
\end{tabular}
\end{sc}
\end{small}
\end{center}
\vskip -0.1in
\end{table*}

\begin{table*}[t]
\caption{Average number of activated experts when using \textbf{simplified OEA} (top-$k_0$ + piggybacking) on Qwen3-235B-A22B.} 
\label{table:oae235b-avg-experts}
\vskip 0.15in
\begin{center}
\begin{small}
\begin{sc}
\begin{tabular}{lccccc}
\toprule
 & $k_0=3$ & $k_0=4$ & $k_0=5$ & $k_0=6$ & vanilla \\
\midrule
aime24 & 27.5 & 32.9 & 38.4 & 43.9 & 53.2 \\
gpqa & 27.4 & 33.3 & 38.6 & 43.1 & 51.6 \\
livecodebench & 28.8 & 34.9 & 41.2 & 45.8 & 55.1 \\
math\_500 & 29.6 & 36.2 & 42.4 & 46.2 & 56.0 \\
\midrule
average & 28.3 & 34.4 & 40.2 & 44.7 & 54.0 \\
normalized average & 0.53 & 0.64 & 0.74 & 0.83 & 1.00 \\
\bottomrule
\end{tabular}
\end{sc}
\end{small}
\end{center}
\vskip -0.1in
\end{table*}

\section{Cross entropy ablation plots}
\subsection{Pruned vs OEA ablation}
We performed ablations for all batch sizes showing the Pareto frontier of OEA-based experiments in contrast to simple pruning (based on phase 1). The results are shown in Figure~\ref{fig:all-pruned-vs-oae} and confirm the piggybacking's (phase 2) gains.
\sweepquad{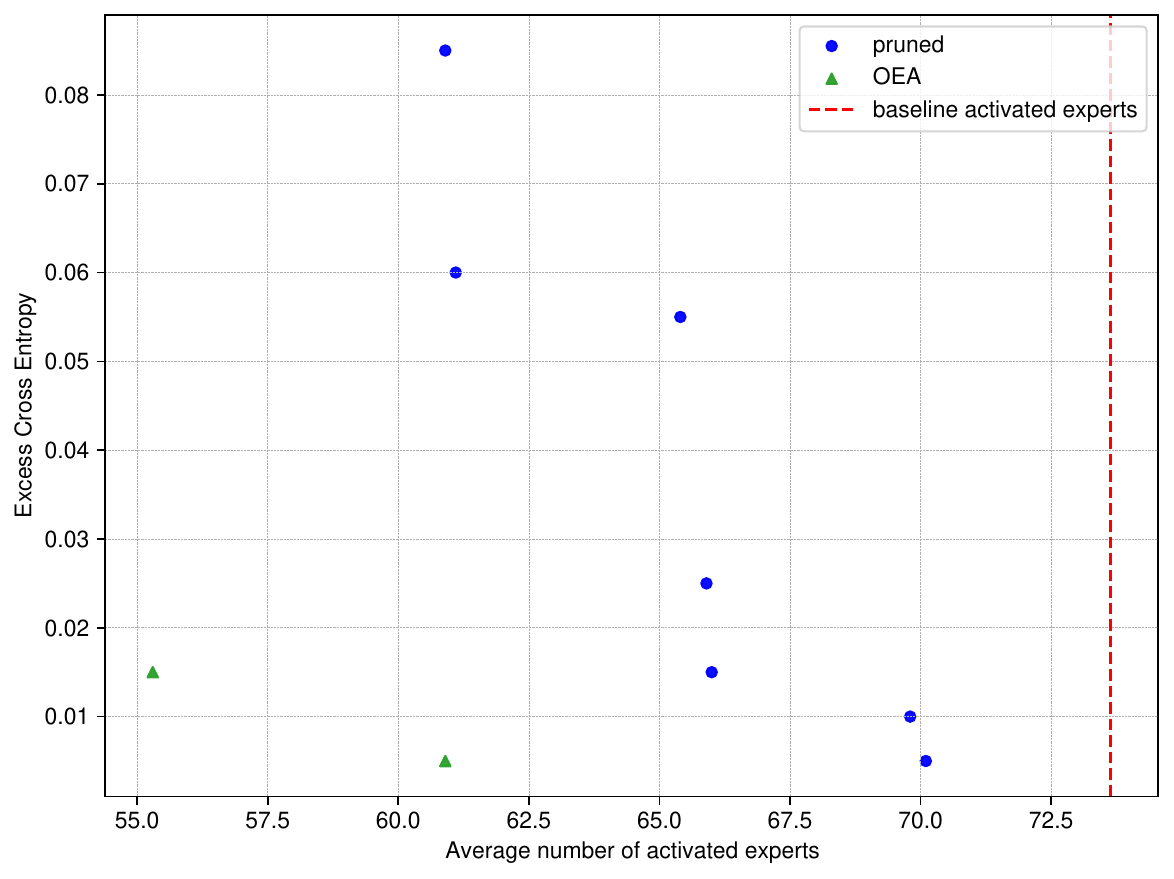}{The y-axis shows the cross-entropy delta relative to the baseline (lower left is better). The two types of dots correspond to the Pareto frontiers of pruned and OEA experiments at all batch-sizes $B$. OEA consistently performs better.}{fig:all-pruned-vs-oae}

\subsection{Ablation over $\text{maxP}$}
Pareto frontiers corresponding to each of the values of $\text{maxP}$ are displayed in Figure~\ref{fig:all-maxp} for all batch sizes $B$ and support the fact that $\text{maxP}=128$ is optimal while $\text{maxP}=8$ is strictly worse.
\sweepquad{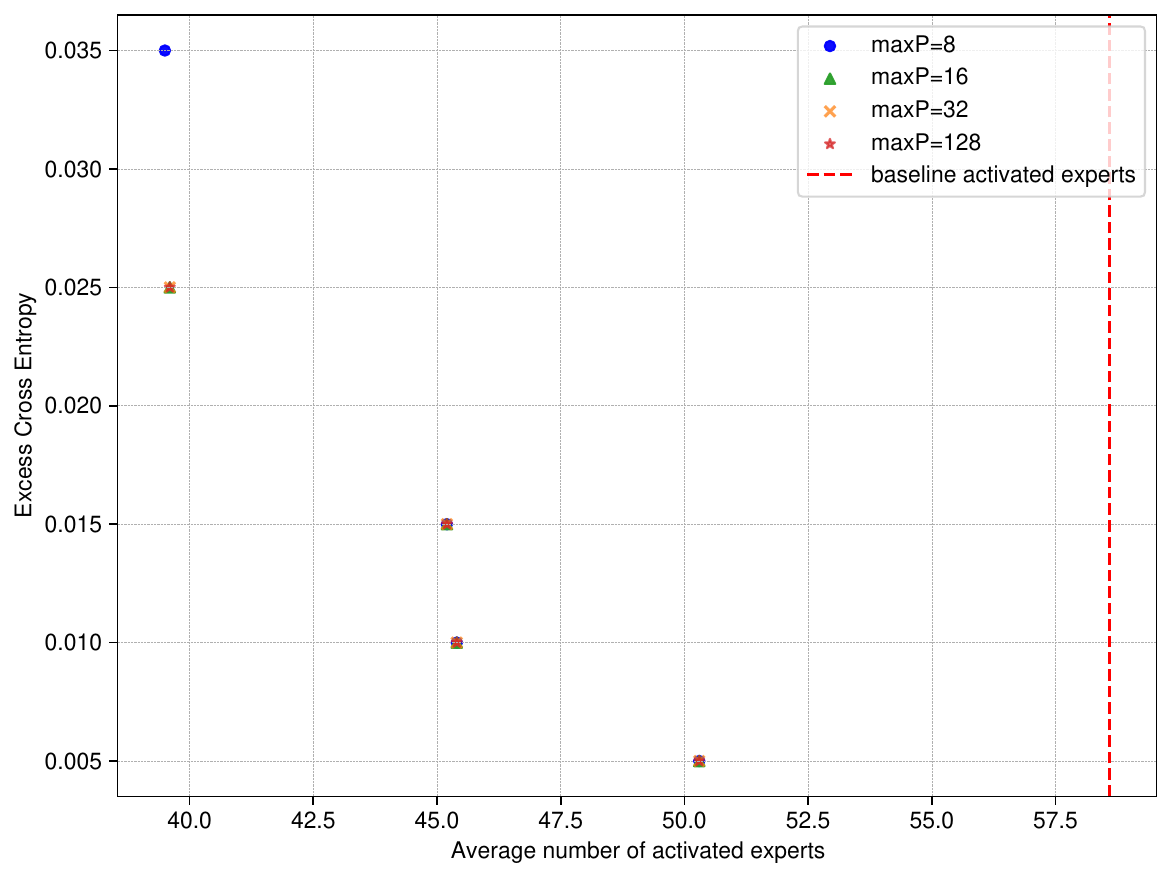}{The y-axis shows the cross-entropy delta relative to the baseline (lower left is better). The four types of dots correspond to the Pareto frontiers of experiments using different values of $\text{maxP}$. $\text{maxP}$ consistently performs best, whereas $\text{maxP}=8$ is strictly worse than it.
}{fig:all-maxp}

\subsection{Ablation over $k^\text{max}$}
Ablations over the value of $k^\text{max}$ are depicted in Figure~\ref{fig:all-maxk}, with a different Pareto frontier computed for each $k^\text{max}$. Note that $k^\text{max}=8$ and $k^\text{max}=9$ perform comparably with others being strictly worse. 
\sweepquad{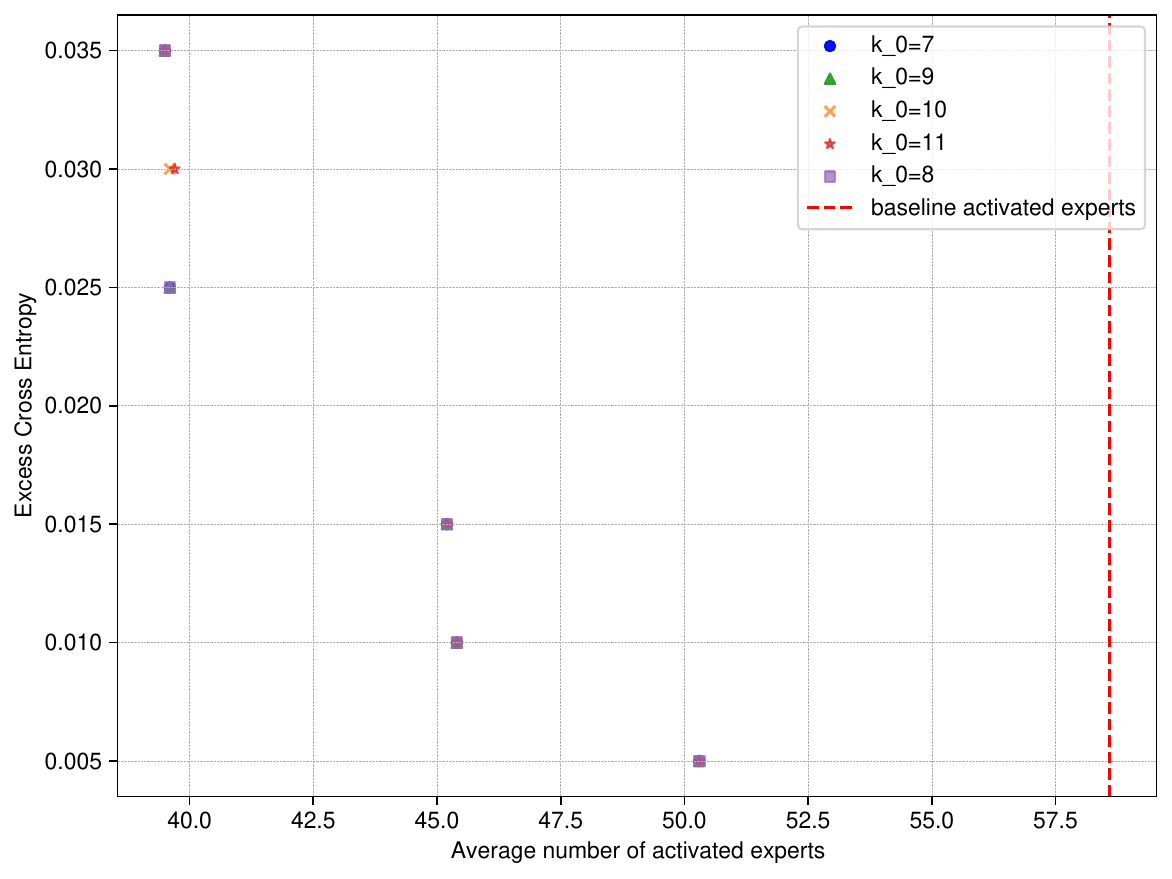}{The y-axis shows the cross-entropy delta relative to the baseline (lower left is better). The five types of dots correspond to the Pareto frontiers experiments using different values of $k^\text{max}$. $k^\text{max}\in\{8,9\}$ perform best while all others perform strictly worse.}{fig:all-maxk}

\subsection{Simplified OEA contrasted with other settings}
Figure~\ref{fig:all-simplified} contrasts the Pareto frontier of simplified OEA (Algorithm~\ref{alg:simplified-router}) with all the other settings (pruned and general OEA together). It shows no meaningful trade-off losses.
\sweepquad{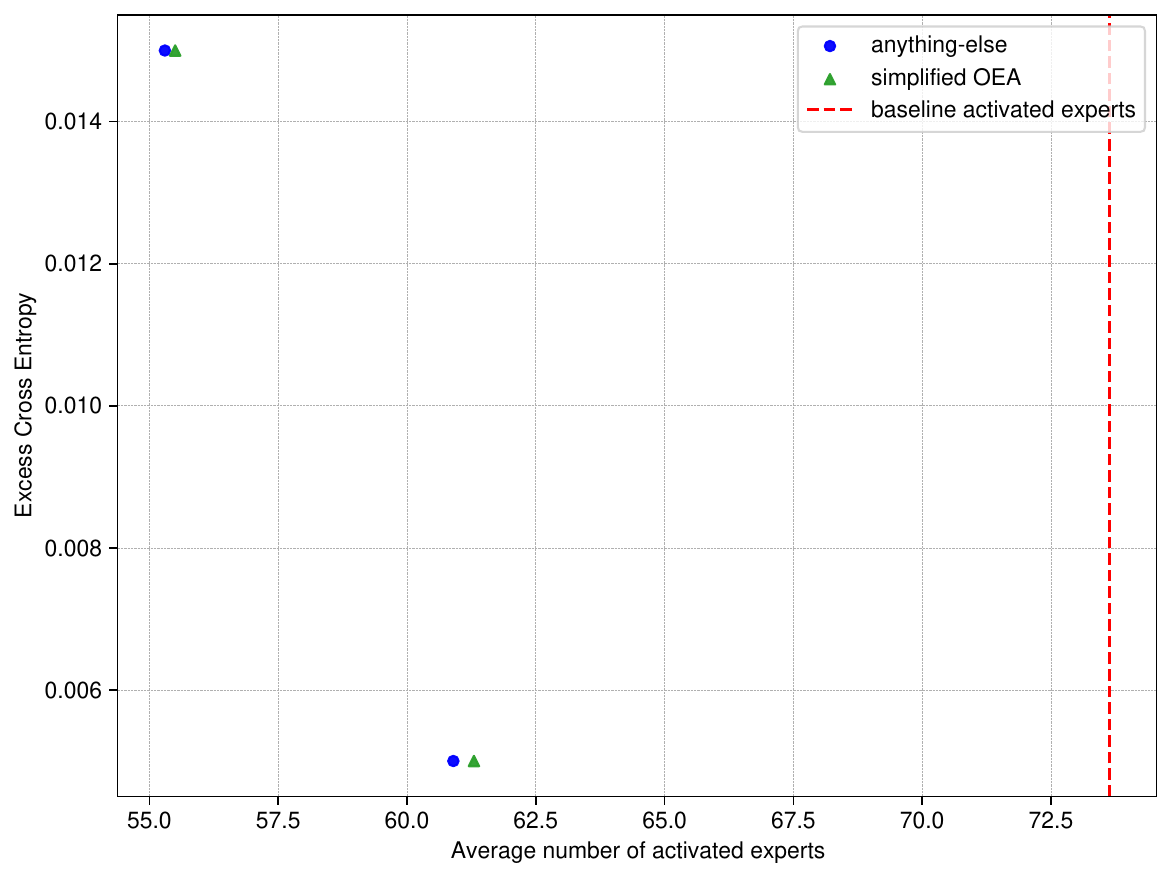}{The y-axis shows the cross-entropy delta relative to the baseline (lower left is better). The two types of dots correspond to the Pareto frontiers of simplified OEA and the rest of experiments at all batch-sizes $B$. Simplified OEA performs comparably to the best hyperparameter choices.}{fig:all-simplified}

\subsection{Ablation over $p$}
We grouped experiments by whether $p=1$ (thus having a static $k_0$ core experts per token) or $p < 1$, as well as whether they use a pruned (phase-1) routing or an OEA-based routing. Pareto frontiers for these $4$ groups are depicted in Figure~\ref{fig:all-p}. Note that within both pruned and OEA, it consistently holds that $p=1$ approximately recovers performance of $p<1$.
\sweepquad{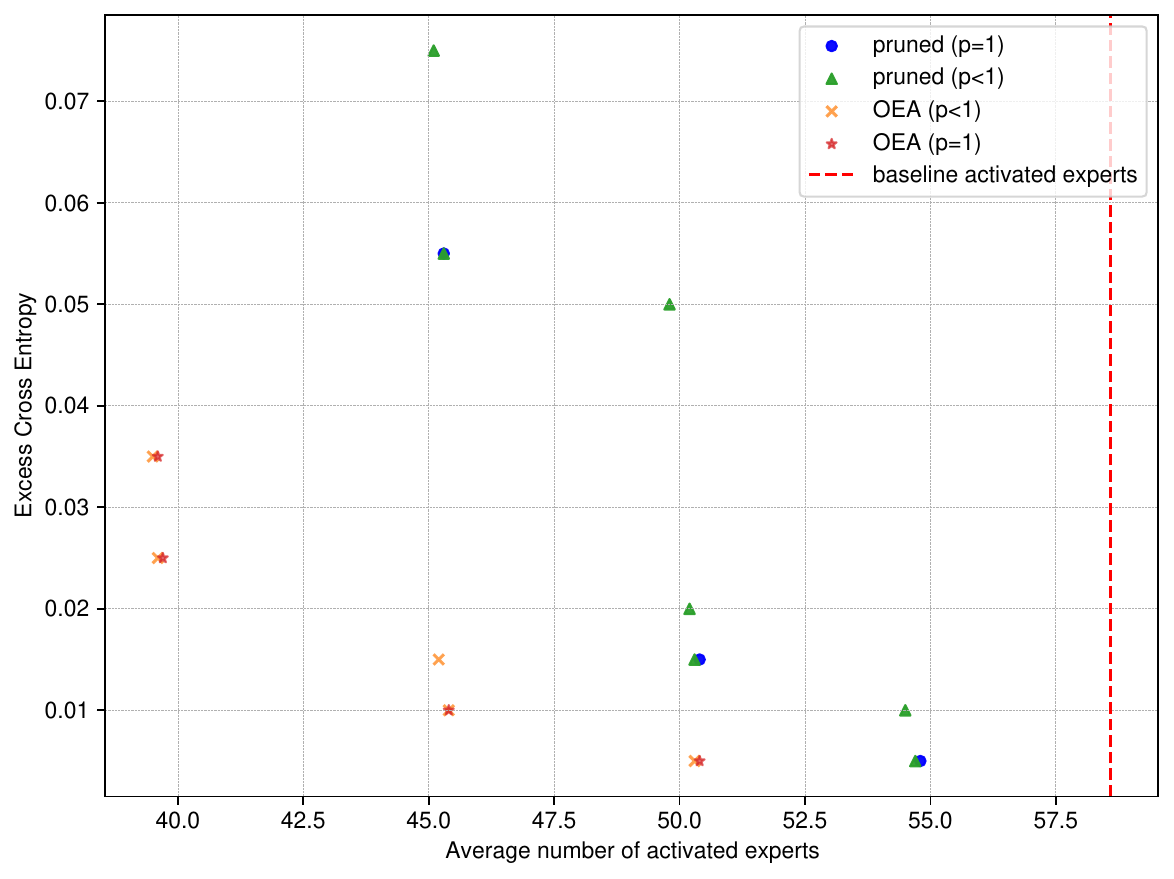}{The y-axis shows the cross-entropy delta relative to the baseline (lower left is better). We split dots as per the legend (by whether $p=1$ and whether they use pruned or OEA-based routings) and report the Pareto frontiers of each group for all batch sizes $B$. Always using $p=1$ does not compromise substantial performance within either group.}{fig:all-p}


\end{document}